%% file: main.tex
\documentclass[runningheads]{llncs}

\usepackage{eccv}

\usepackage{eccvabbrv}

\usepackage{graphicx}
\usepackage{booktabs}
\usepackage{multirow}

\usepackage[accsupp]{axessibility}  

\usepackage{hyperref}

\usepackage{orcidlink}

 \NewDocumentCommand{\suzie}{ mO{} }{\textcolor{cyan}{\textsuperscript{\textit{Suzie}}\textsf{\textbf{\small[#1]}}}}

 \NewDocumentCommand{\kons}{ mO{} }{\textcolor{orange}{\textsuperscript{\textit{Kons}}\textsf{\textbf{\small[#1]}}}}
 
 \NewDocumentCommand{\luca}{ mO{} }{\textcolor{olive}{\textsuperscript{\textit{Luca}}\textsf{\textbf{\small[#1]}}}}

 \NewDocumentCommand{\paolo}{ mO{} }{\textcolor{purple}{\textsuperscript{\textit{Paolo}}\textsf{\textbf{\small[#1]}}}}

 \NewDocumentCommand{\FG}{ mO{} }{\textcolor{green}{\textsuperscript{\textit{FG}}\textsf{\textbf{\small[#1]}}}}

\begin{document}

\title{Hyperbolic Learning with Multimodal Large Language Models} 


\author{Paolo Mandica\thanks{Equal contribution}\inst{1}\orcidlink{0000-0002-4493-2497} \and
Luca Franco\inst{\star 2}\orcidlink{0000-0003-0107-6755} \and
Konstantinos Kallidromitis\inst{3} \and \\
Suzanne Petryk\inst{4} \and
Fabio Galasso \inst{1}\orcidlink{0000-0003-1875-7813}
}

\authorrunning{P.~Mandica et al.}

\institute{Sapienza University of Rome \and
ItalAI s.r.l.\\
\url{https://italailabs.com/} \and
Panasonic Corp. of North America \and
University of California Berkeley}

\maketitle

\begin{abstract}

Hyperbolic embeddings have demonstrated their effectiveness in capturing measures of uncertainty and hierarchical relationships across various deep-learning tasks, including image segmentation and active learning. However, their application in modern vision-language models (VLMs) has been limited. A notable exception is MERU, which leverages the hierarchical properties of hyperbolic space in the CLIP ViT-large model, consisting of hundreds of millions parameters.
In our work, we address the challenges of scaling multi-modal hyperbolic models by orders of magnitude in terms of parameters (billions) and training complexity using the BLIP-2 architecture.
Although hyperbolic embeddings offer potential insights into uncertainty not present in Euclidean embeddings, our analysis reveals that scaling these models is particularly difficult. 
We propose a novel training strategy for a hyperbolic version of BLIP-2, which allows to achieve comparable performance to its Euclidean counterpart, while maintaining stability throughout the training process and showing a meaningful indication of uncertainty with each embedding.

\end{abstract}

\input{sections/introduction}

\input{sections/related}

\input{sections/background}

\input{sections/method}

\input{sections/experiments}

\input{sections/discussion}

\input{sections/acknowledgements}

\bibliographystyle{splncs04}
\bibliography{main}
\end{document}

%% file: sections/introduction.tex
\section{Introduction}
\label{sec:intro}
To harness the recent advancements in language models across different modalities, modern vision-language models (VLMs) have evolved to combine visual processing with the reasoning capabilities of large language models (LLMs)\cite{li2023blip,liu2024visual,alayrac2022flamingo,yin2023survey}. These integrated architectures enable reasoning tasks and encompass extensive world knowledge, facilitating capabilities such as zero-shot segmentation and improved generalization for images \cite{lai2024lisa}. Models like BLIP-2 (Bootstrapping Language-Image Pre-training) combine language understanding with visual inputs through a two-stage pre-training approach \cite{li2023blip}. In the first stage, BLIP-2 aligns visual features with textual descriptions. In the second stage, it fine-tunes this alignment for tasks such as image captioning and visual question answering. Despite these advancements, BLIP-2 still faces limitations, particularly in representing complex hierarchical structures and relationships in the data. These limitations are crucial during image and text alignment and highlight the need for an uncertainty measure to ensure consistency between image and language representations.

To better understand the embedding space of an LLM, we compare its Euclidean and hyperbolic counterparts. Hyperbolic space provides a more efficient and compact representation of hierarchical structures due to its exponential scaling properties \cite{desai2023hyperbolic, mettes2024hyperbolic}. In contrast, Euclidean space grows linearly, making it challenging to represent hierarchical data with low distortion. Leveraging the hyperbolic space allows models to achieve an effective understanding of hierarchical data, leading to improved performance in tasks such as active learning \cite{franco2023hyperbolic}, action recognition \cite{franco2023hyperbolic2}, and segmentation \cite{atigh2022hyperbolic}. Additionally, hyperbolic embeddings contain an uncertainty measure represented by the distance from the center of the Poincaré ball, a specific instance of a hyperbolic manifold. However, it has not yet been mathematically proven how this property is achieved. Some works \cite{chen2023hyperbolic, atigh2022hyperbolic, suris2021hyperfuture} show that a lower radius is associated with higher uncertainty, and conversely at a high radius there are less uncertain samples. In contrast, other recent work \cite{franco2023hyperbolic} links radius to complexity or specifically epistemic uncertainty. In this case the situation is reversed; at low radius there are more generic and common classes, while at high radius more complex and specific classes. Although some work is still needed in this direction, utilizing this distance-based measure of uncertainty models can better assess and manage the confidence levels of their predictions, leading to a more reliable and interpretable grounding of vision and language.

Although some studies \cite{desai2023hyperbolic, ibrahimiintriguing} have explored the use of hyperbolic space in medium-sized vision-language models like CLIP \cite{radford2021learning}, hyperbolic embeddings remain largely unexplored in large-scale modern vision-language models (VLMs) such as BLIP-2, which features 2.7 billion parameters and maps vision-based features into a shared multimodal space with text before passing them to the LLM. We propose combining BLIP-2 with hyperbolic embeddings to enhance its representational power, improving the model's ability to capture intricate relationships and perform complex reasoning, while also enabling the use of hierarchical data structures in multimodal learning. In this work, we demonstrate that traditional hyperbolic approaches, such as using Poincaré distance as a similarity measure, fail when applied within BLIP-2. To overcome these limitations, we propose a stable hyperbolic training method that maintains performance with the Euclidean metrics on retrieval tasks while producing image embeddings with varying hyperbolic radii. Our key contributions are as follows:
\begin{itemize}
    \item We present, to our knowledge, the first large-scale hyperbolic VLM and show that hyperbolic embeddings can capture uncertainty information via their radii, while achieving performance comparable to the Euclidean baseline of BLIP-2;
    \item We provide detailed quantitative and visual analyses of the different embedding spaces, offering insights into the operation of current VLMs and how to efficiently represent them in hyperbolic space.
\end{itemize}

%% file: sections/related.tex
\section{Related Works}
\label{sec:related}

\subsection{Hyperbolic Representation Learning}
\label{sec:hrl}
Hyperbolic representation learning in deep neural networks gained momentum with \cite{ganea2018hyperbolic}, introducing hyperbolic counterparts for classical fully-connected layers, multinomial logistic regression, and RNNs. This approach later expanded to hyperbolic convolutional neural networks \cite{shimizu2020hyperbolic}, hyperbolic graph neural networks \cite{liu2019hyperbolic, chami2019hyperbolic}, and hyperbolic attention networks \cite{gulcehre2018hyperbolic}. Hyperbolic geometry's ability to encode hierarchies and tree-like structures \cite{nickel2017poincare, tifrea2018poincar, chami2019hyperbolic, khrulkov2020hyperbolic, suris2021hyperfuture, atigh2022hyperbolic} as well as the notion of uncertainty via embedding radius \cite{ermolov2022hyperbolic, chen2023hyperbolic, flaborea23, franco2023hyperbolic2} have been key motivations. Recent advancements include self-supervised learning in hyperbolic space \cite{yan2021unsupervised, suris2021hyperfuture, ermolov2022hyperbolic, franco2023hyperbolic, franco2023hyperbolic2}, where hierarchy or uncertainty measure emerge without a direct supervision.

Transposing models from Euclidean to hyperbolic is challenging due to the specific hyperparameter configurations required, particularly the curvature of the manifold, which is data-dependent \cite{khrulkov2020hyperbolic, ermolov2022hyperbolic}. Hyperbolic neural networks have been explored for encoding full images \cite{Spengler2023PoincarR, Bdeir2024}, single pixels \cite{atigh2022hyperbolic, franco2023hyperbolic, chen2023hyperbolic}, and text \cite{tifrea2018poincar, Dhingra2018EmbeddingTI}. Combining different modalities is even more complex. \cite{desai2023hyperbolic} introduced MERU, a CLIP-like model to align text and image modalities. However, these models are still not comparable to large-scale and multi-purpose models like BLIP-2 \cite{li2023blip}, which leverage extensive pre-training and fine-tuning for tasks such as VQA and captioning.


\subsection{Image-Text Alignment}
\label{sec:itrl}
Mapping images and text into the same latent space has been a useful technique within a range of multimodal tasks, such as cross-modal retrieval.
It is most directly useful for cross-modal retrieval tasks. CLIP~\cite{radford2021learning} is a popular model in this space, consisting of image and text encoders trained on large-scale data to embed images and text into a shared latent space, using contrastive learning. Recent VLMs have taken a different approach, training multimodal adapter layers between a vision encoder and LLM~\cite{instructblip,li2023blip,liu2024visual,ye2023mplug,zhu2023minigpt}. This aligns visual representations with the LLM input space. As part of training the multimodal adapter, BLIP-2 and InstructBLIP~\cite{instructblip,li2023blip} additionally use an image-text contrastive loss similar to CLIP. 

In our work, we explore a hyperbolic version of BLIP-2, presenting our findings and highlighting the challenges in leveraging contrastive learning for hyperbolic representations. 

%% file: sections/background.tex
\section{Background}
\label{sec:background}

\subsection{Hyperbolic Geometry}
Hyperbolic neural networks operate within hyperbolic space, characterized by a constant negative curvature. A commonly employed manifold in this context is the Poincaré ball, selected due to its favorable geometric properties. The Poincaré ball is defined as \((\mathbb{D}_c^N, g^{\mathbb{D}_c})\), with:
\begin{equation}
    \mathbb{D}_c^N = \{x \in \mathbb{R}^N : c \|x\| < 1 \}
\label{eq:poincare_ball}
\end{equation}
where \(c\) is the curvature and \(\|\cdot\|\) denotes the standard Euclidean \(L2\)-norm. The associated Riemannian metric \(g_x^{\mathbb{D}_c}\) is given by:
\begin{equation}
    g_x^{\mathbb{D}_c} = (\lambda_x^c)^2 g^\mathbb{E}
\end{equation}
where \(\lambda_x^c = \frac{2}{1 - c\|x\|^2}\) is the conformal factor and \(g^\mathbb{E} = \mathbb{I}^N\) is the Euclidean metric tensor.

\subsubsection{Exponential Map and Möbius Addition}
In hyperbolic neural networks, a feature vector $v\in \mathbb{R}^N$ is firstly extracted in Euclidean space, and then it is projected into the Poincaré ball using the exponential map:
\begin{equation}
    \text{exp}_x^c(v) = x \oplus_c \left(\frac{v}{\sqrt{c}\|v\|} \tanh\left(\sqrt{c} \frac{\lambda_x^c \|v\|}{2}\right)\right)
\label{eq:expmap}
\end{equation}
where \(x \in \mathbb{D}_c^N\) is the anchor, and \(\oplus_c\) denotes the Möbius hyperbolic addition, defined for two hyperbolic vectors \(h, w\) as:
\begin{equation}
    h \oplus_c w = \frac{(1 + 2c \langle h, w \rangle + c \|w\|^2)h + (1 - c \|h\|^2)w}{1 + 2c \langle h, w \rangle + c^2 \|h\|^2 \|w\|^2}
\label{eq:mobius_add}
\end{equation}

\subsubsection{Poincaré Distance and Hyperbolic Radius}
The Poincaré distance \(d_{Poin}\) between two hyperbolic vectors \(x, y \in \mathbb{D}^N_c\) is defined as:
\begin{equation}
    d_{Poin}(x, y) = \frac{2}{\sqrt{c}} \tanh^{-1} \left(\sqrt{c} \| -x \oplus_c y \|\right)
\label{eq:poincare_dist}
\end{equation}
We define the hyperbolic radius of the embedding $h \in \mathbb{D}_c^N $ as the Poincaré distance of $h$ from the origin of the ball $d_{Poin}(0, h)$.

\subsection{BLIP-2}
In BLIP-2, Li et al. \cite{li2023blip} leverage pre-trained vision and language models to reach a strong multimodal understanding with minimal additional training. More specifically, it begins by extracting a visual embedding from an image $I$ using a pre-trained and frozen image encoder $M_{\text{image}}$. The embedding for image $I$ is thus computed as:

\begin{equation}
    v = M_{\text{image}}(I)
\label{eq:blip2-vision}
\end{equation}

Next, a relatively small adapter model, called a Q-Former (or Querying-Transformer) $M_{\text{QF}}$, is trained along with a projection layer $P$ to translate the embedding into the input space of a pre-trained and frozen LLM $M_{\text{text}}$ (e.g., OPT~\cite{zhang2022opt}). The Q-Former is a transformer encoder that takes in a sequence of $N$ fixed ``query'' vectors $\mathbf{Q} = \left[q^j \right]_{j=1}^N$ that interact with the image embedding $v$ via cross-attention.
This produces output query embeddings $\mathbf{E}$:

\begin{equation}
    \mathbf{E} = \left[e^j \right]_{j=1}^N = M_{\text{QF}}(v, \mathbf{Q})
\label{eq:blip2-adapter}
\end{equation}

In parallel, a text encoder $M_{\text{text}}$ represents the caption $T$ relative to the image. 
\begin{equation}
    u = M_{\text{text}}(T)
\label{eq:blip2-text}
\end{equation}
For clarity, we will consider the text encoder $M_{\text{text}}$ as part of the Q-former.

In BLIP-2, the text is aligned via cosine distance with the most similar query embedding in $\mathbf{E}$.

\begin{equation}
    L_{\text{contr}}(\mathbf{E}, u) = \min_{j=1,...,N} d_{cosine}(e^j, u) = \min_{j=1,...,N} 1 - \frac{e^j \cdot u}{\|e^j\| \|u\|}
\label{eq:blip2-contr}
\end{equation}

Finally, each embedding $e^j$ is projected by the linear layer $P$ to match the dimensionality of the LLM input token embeddings. To generate text, the LLM prompt begins with the sequence of embeddings $\mathbf{E}$, followed by the rest of the tokenized prompt.

%% file: sections/method.tex
\section{Stabilizing Hyperbolic BLIP-2}
\label{sec:method}



In this section, we discuss challenges encountered in adopting standard approaches to hyperbolic contrastive learning and the solutions we implemented. Additionally, we introduce two novel strategies: Random Query Selection (RQS) and Random Text Pruning (RTP). These strategies aim to effectively utilize hyperbolic embeddings while enhancing model stability and performance in multimodal tasks.

\subsection{Hyperbolic Image-Text Contrastive}
\label{subsec:hyper_contrastive}
We applied contrastive learning in hyperbolic space by first mapping the Euclidean feature vectors $u$ and $e^j$ relative to the text and image inputs into the Poincaré ball using exponential mapping (see Eq. \ref{eq:expmap}):

\begin{equation}
    u_h = \text{exp}_0^c(u), \quad e_h^j = \text{exp}_0^c(e^j) \quad \text{for} \ j=1,...,N
\end{equation}

This transformation positions the embeddings $u_h, \left[e_h^j \right]_{j=1}^N$ within the hyperbolic space, preserving their relational structure.

Next, we substitute the distance used in \ref{eq:blip2-contr} with the hyperbolic metric, specifically the Poincaré distance (see Eq. \ref{eq:poincare_dist}):

\begin{equation}
    L_{\text{hyper}}(\mathbf{E_h}, u_h) = \min_{j=1,...,N} d_{Poin}(e_h^j, u_h)
\label{eq:hyper-contr}
\end{equation}

This distance measure, based on hyperbolic geometry, calculates the length of the geodesic between two embeddings. Notably, embeddings closer to the edge of the Poincaré ball result in longer geodesics and thus greater distances.

Following the approach in \cite{Ge2022HyperbolicCL}, we utilized the Poincaré distance to assess the similarity between embeddings. However, as noted in \cite{franco2023hyperbolic2}, the Poincaré distance is effective for positive-only samples \cite{byol, Chen2020ExploringSS} but introduces ambiguity when applied to both positive and negative samples \cite{simclr}. This ambiguity arises because repelling negative samples along geodesics can push embeddings toward the edge of the ball, thereby biasing the uncertainty.

Given these considerations and our experimental results, we opted to use the standard cosine distance with hyperbolic image and text embeddings:

\begin{equation}
    L_{\text{hyper-cosine}}(\mathbf{E_h}, u_h) = \min_{j=1,...,N} d_{cosine}(e_h^j, u_h)
\label{eq:hyper-contr-cosine}
\end{equation}

This choice aims to preserve the performance of the Euclidean baseline while exploring whether the model can still learn uncertainty effectively within the hyperbolic space.

\subsection{Random Query Selection}
\label{subsec:rqs}

The baseline model utilizes the Q-Former architecture \cite{li2023blip}, which outputs 32 learned query embeddings $\mathbf{E_h}$. These embeddings are designed to capture diverse semantic information from the image and align it with text. However, when computing the contrastive learning loss (e.g., Eq. \ref{eq:blip2-contr}, \ref{eq:hyper-contr}, \ref{eq:hyper-contr-cosine}), the model selects the most similar query embedding exclusively. This approach limits the utilization of the rich information stored in the remaining query tokens. As a result, the model tends to merge the different semantic information into a single query embedding, consistently selecting the same embedding for all inputs (c.f., Fig. \ref{fig:rand_query_plot}).

To address this limitation, we introduce Random Query Selection (RQS) to diversify the semantics captured by the different query embeddings. Specifically, during training, the model selects the most similar query embedding to the reference text embedding with a probability $p$. With a probability $1 - p$, the model selects a random query embedding instead:

\begin{equation}
    L_{\text{hyper-RQS}}(\mathbf{E_h}, u_h) = 
    \left\{
    \begin{array}{ll}
      \min_j d_{cosine}(e_h^j, u_h), & \text{with prob}=p\\
      d_{cosine}(e_h^{\tilde{j}}, u_h), & \text{with prob}=1-p
    \end{array}
    \right.
\label{eq:hyper-RQS}
\end{equation}

where $\tilde{j}$ is uniformly sampled in $\{1, ..., N\}$. This stochastic selection process encourages the model to distribute the semantic information across the 32 query embeddings, thereby enhancing the robustness and diversity of the learned representations.

\subsection{Random Text Pruning}
\label{subsec:rtp}

In our model, the hyperbolic embeddings $u_h$ associated with text inputs tend to migrate towards the edge of the Poincaré ball. This results in uniform radii, rendering them ineffective as a proxy for uncertainty (c.f. Fig. \ref{fig:hyper_radii_different_dimension} right in purple). To address this, we induce meaningful variations in text radii by introducing noise into the text inputs. This is expected to reflect different uncertainty levels through varying noise.

To achieve this, we implement Random Text Pruning (RTP), where a randomly selected segment of the input text \( T \) is pruned. Specifically, a window of 0 to 7 consecutive tokens is removed, introducing varying levels of noise. Through experimentation, we determined that a window length of 7 yields the best performance.

By applying RTP, we expect the embeddings of noisier texts to exhibit lower radii, aligning with the interpretation of hyperbolic representation where increased uncertainty corresponds to a reduced radius. This technique encourages the model to produce hyperbolic embeddings with radii that reflect the underlying uncertainty of the text inputs, thereby enhancing the expressiveness and utility of the hyperbolic space in representing text semantics.

%% file: sections/experiments.tex
\section{Experiments}
\label{sec:experiments}

This section outlines the training pipeline for our models, analyzes the learned hyperbolic embeddings, details the evaluation metrics, and presents the results on downstream tasks.

\subsection{Training Pipeline}

The training pipeline follows the methodology described in BLIP-2, integrating vision-language pre-training with contrastive learning in hyperbolic space. The main components of the training pipeline are outlined below.

\paragraph{Dataset}
The trainings of our models were conducted on the COCO dataset \cite{cocodataset}. \textbf{COCO (Common Objects in Context)} consists of over 200,000 images with a  wide range of annotations for object detection, segmentation, and captioning. It captures a variety of everyday scenes with common objects in their natural contexts.

\paragraph{Model Architecture:}
Our model architecture includes a frozen pre-trained state-of-the-art vision transformer, ViT-g/14 from CLIP \cite{radford2021learning}, a frozen language model OPT \cite{zhang2022opt}, and the Q-former. The outputs of the vision and language encoders are projected into the Poincaré ball for contrastive learning in the Q-former.

\paragraph{Model Training:}
The model undergoes a two-stage pre-training process: 250k steps in the first stage and 80k steps in the second, following the same setup as BLIP-2. The batch size is 800 for the first stage and 512 for the second. Each image is resized to 224x224 with standard BLIP-2 augmentations. Pre-training is conducted on a single machine with 8 V100 (32GB) GPUs. Stage 1 takes 12 hours, while Stage 2 takes 10 hours. Both stages use the same set of hyperparameters: AdamW optimizer with $\beta_1 = 0.9$, $\beta_2 = 0.98$, and a weight decay of 0.05. The learning rate starts at 1e-6 with a 2000-step warmup to 1e-4, then decays following a cosine schedule to a minimum of 1e-5.
After pre-training, the model is fine-tuned on the image captioning task.

\subsection{Analysis}

This section systematically examines the principal findings derived from our experiments and visualizations. Our objective was to establish the radius as an indicator of uncertainty. We provide a detailed analysis of the results obtained from employing conventional methodologies for hyperbolic representations, as mentioned in \ref{subsec:hyper_contrastive}. Furthermore, we demonstrate the challenges encountered during this process and the corresponding countermeasures implemented to mitigate these issues.

\input{figures/scripts/image_radii_hyper}

\paragraph{\textbf{Hyperbolic Radius as a Measure of Uncertainty }}

In Figure \ref{fig:image_radii_hyper}, we visualize how the hyperbolic radius provides insights into its potential as a measure of uncertainty or complexity in the embeddings. Our experiments revealed that the radius can exhibit variability, hinting at the degree of uncertainty or the complexity of the embedded information. In fact, at a lower radius (see Figure \ref{fig:image_radii_hyper} top), we observe "people skiing" where the majority of the image is dominated by background classes like snow, sky, or vegetation, while at a higher radius (see Figure \ref{fig:image_radii_hyper} bottom), we have more complex scenarios with tables full of different objects.

In stark contrast to previous works such as\cite{franco2023hyperbolic2}, we observe that the average radius distribution of the different classes \footnote{The classes are the ones provided in the COCO dataset that are used just for the analysis and comparison in the literature.} (c.f. Figure \ref{fig:radii_distribution}) is mainly uniform for intermediate radii, while it looks significant only at the extremes of the radii distribution.

\input{figures/scripts/radii_distribution}

\paragraph{\textbf{Challenges in Contrastive Learning with Hyperbolic Space}}

Following the observations from \cite{franco2023hyperbolic2}, we found that applying contrastive learning in hyperbolic space can be problematic when utilizing both positive and negative samples, as in \cite{simclr}. The Poincaré distance, while effective for measuring geodesic lengths in hyperbolic space, introduces ambiguity when repulsing with negative samples, potentially biasing the embeddings towards the edge of the ball. This issue does not arise with positive-only samples \cite{byol}, suggesting a more stable training dynamic in those scenarios.

Another critical challenge identified was the difficulty of fine-tuning the Q-former, pre-trained in the Euclidean manifold and then frozen, within hyperbolic space. The misalignment between the pre-trained Euclidean image encoder and the hyperbolic representation space caused significant performance issues, indicating the necessity for consistency in embedding space across training stages.

\paragraph{\textbf{Hyperbolic Mapping and Similarity Measures}}

Our findings in Table \ref{tab:main_results}, indicate that hyperbolic mapping remains effective only when the dot product is used to compute similarity. When the Poincaré distance is employed, the performance degrades, rendering it ineffective for the contrastive loss. However, similarities derived from the Poincaré distance can be useful for negative sample selection in Image-Text Matching tasks, providing a nuanced approach to leveraging hyperbolic geometry.
\input{figures/scripts/hyper_radii_different_dimension}

\paragraph{\textbf{Impact of Embedding Dimension on Radius and Performance}}

In Figure \ref{fig:hyper_radii_different_dimension} (cyan plot), we observe that the hyperbolic radius tends to cap at its maximum level (the edge of the Poincaré ball) when using high-dimensional embeddings (i.e., 256). Consequently, the radius does not affect the training, and we do not have a meaningful radius as an uncertainty proxy. We see a significant variation in the radius by reducing the embedding dimension (purple plot) to 16. Performance metrics show a notable decline when the embedding dimension was reduced to lower values (e.g., 16), while higher dimensions (e.g., 128) maintained comparable performance to Euclidean models. These results highlight the sensitivity of hyperbolic embeddings to dimensionality, affecting both the interpretability of the radius and the overall model performance.
\paragraph{\textbf{Variability of Image and Text Embeddings' Radii}}

Figure \ref{fig:hyper_radii_different_dimension} shows a distinct behavior between image (left) and text embeddings (right). Image embeddings exhibit variable radii, whereas text embeddings consistently stay at the edge of the Poincaré ball. 
Feature clipping \cite{Guo2021ClippedHC} is used in the current literature to help in training stabilization and to alleviate this effect. However, we do not observe any significant influence on the performance of the hyperbolic model or the hyperbolic radius.
For this reason, we introduce noise through Random Text Pruning (c.f. Sec. \ref{subsec:rtp}), which causes slight variations in the text embeddings' radii (green plot), indicating a potential but limited ability to capture text uncertainty through hyperbolic geometry.

\input{figures/scripts/rand_query_plot}

\paragraph{\textbf{Random Query Selection}}

The baseline model employs the Q-former architecture, which generates 32 learned query embeddings. These embeddings are intended to encapsulate diverse semantic information from the image and align it with text semantics. However, when calculating the contrastive learning loss with the text embedding, the model exclusively selects the most similar query embedding.

In Figure \ref{fig:rand_query_plot} (top), we present the distribution of the selected query embeddings for the baseline BLIP-2 model. Our observations indicate that query 31 is almost invariably chosen as the embedding that best represents the image. By introducing random query selection, we achieve a more distributed representation of information (see Figure \ref{fig:rand_query_plot} bottom). In stage 2, when the LLM receives all the query embeddings as input, it benefits from more informative embeddings that capture various aspects of the input image, thereby enhancing performance.

\subsection{Evaluation Metrics}
The performance of our model is evaluated using standard metrics for the downstream task of zero-shot image-text retrieval and image captioning. 

\paragraph{Image Retrieval @1} measures the percentage of times the correct image is retrieved as the top result when given a text query.

\paragraph{Text Retrieval @1} measures the percentage of times the correct text is retrieved as the top result when given an image query.

\paragraph{BLEU (Bilingual Evaluation Understudy)} \cite{bleu2002} measures the precision of n-grams in the generated text that match n-grams in the reference text, and it is widely used for its strong correlation with human judgment.

\paragraph{METEOR (Metric for Evaluation of Translation with Explicit ORdering)} \cite{meteor2005} calculates a score based on the harmonic mean of unigram precision and recall, incorporating stemming and synonymy matching to correlate well with human judgment at the sentence level.

\paragraph{CIDEr (Consensus-based Image Description Evaluation)} \cite{vedantam2015cider} evaluates image captions by measuring the similarity of a generated sentence to a set of ground truth sentences, emphasizing consensus among multiple references.

\paragraph{ROUGE-L (Recall-Oriented Understudy for Gisting Evaluation)} \cite{rouge2004} uses the Longest Common Subsequence (LCS) between the generated text and the reference text to evaluate sentence-level structure similarity.

\paragraph{SPICE (Semantic Propositional Image Caption Evaluation)} \cite{anderson2016spice} compares the semantic propositional content of the generated text with the reference text, focusing on the meaning conveyed by the captions for a nuanced assessment of semantic accuracy.

\subsection{Experimental Results}

To evaluate the contribution of various components in our model, we conduct a series of experiments. Specifically, we compare the performance of the Euclidean and hyperbolic baselines with models that integrate Random Query Selection (RQS) and Random Text Pruning (RTP), as described in Sec. \ref{sec:method}. Additionally, we provide the results of the hyperbolic baseline model using the Poincaré distance in contrastive learning. For the COCO zero-shot retrieval evaluation, we use the stage-1 pre-trained models, while for the image captioning evaluation, we employ the stage-2 models after fine-tuning them for the captioning downstream task. All models utilize a ViT-g vision encoder and, in the image captioning task, the OPT 2.7B language model. The results of these studies are summarized in Table \ref{tab:main_results}. Below, we describe each setup and discuss its impact on the performance metrics.

\input{tables/main_results}

\paragraph{Euclidean Baseline}
The Euclidean baseline model utilizes the BLIP-2 configuration without any hyperbolic modifications. This model serves as a reference point, providing performance benchmarks for both retrieval and captioning tasks. It achieves solid performance in both zero-shot retrieval, with 69.6 for TR@1 and 53.5 for @ IR@1, and image captioning, with scores of 68.6 for BLEU-1, 30.1 for BLEU-4, 29.2 for METEOR, 54.0 for ROUGE-L, 110.7 for CIDEr, and 21.8 for SPICE.

\paragraph{Euclidean RQS}
This model incorporates Random Query Selection (RQS) into the Euclidean baseline. The results demonstrate improvements in most metrics, with a notable increase to 72.1 (+2.5) for TR@1, 55.6 (+2.1) for IR@1, 77.3 (+8.7) for BLUE-1, 36.8 (+6.7) for BLUE-4, and 127.5 (+16.8) for CIDEr, highlighting the benefit of this approach.

\paragraph{Euclidean RQS+RTP}
This model combines Euclidean Random Query selection (RQS) and Text Pruning (RTP) within the BLIP-2 baseline. The inclusion of RTP in combination with RQS further enhances performance, with scores such as 78.8 (+1.5) for BLEU-1, 37.7 (+0.9) for BLEU-4, 58.8 (+0.9) for ROUGE-L, and 129.8 (+2.3) for CIDEr, underscoring the effectiveness of RQS and RTP in the Euclidean space.

\paragraph{Hyperbolic Baseline}
The hyperbolic baseline model serves as the counterpart to the Euclidean baseline but operates within hyperbolic space. This setup allows us to directly compare the effectiveness of hyperbolic versus Euclidean representations in both retrieval and captioning tasks. The performance are slightly lower than the Euclidean counterpart. This is most probably due to the Euclidean pre-trained weights used to initialize the ViT encoder.

\paragraph{Hyperbolic baseline w/ Poincaré distance}
This version of the hyperbolic baseline model uses the Poincaré distance in the computation of the contrastive loss. This model shows a significant drop in performance across all metrics, with scores such as 30.0 for TR@1, 14.0 for BLEU-1, and 0.1 for CIDEr, indicating that using the Poincaré distance with a model pre-trained in Euclidean space creates too much instability.

\paragraph{Hyperbolic RQS}
In this ablation, Random Query selection (RQS) is applied at training time using the hyperbolic model. The results show improvements compared to the hyperbolic baseline, with scores of 71.3 for TR@1, 54.2 for IR@1, and 128.8 for CIDEr, demonstrating the benefits of RQS in hyperbolic space and shortening the gap with the Euclidean RQS model.

\paragraph{Hyperbolic RTP}
This model incorporates Random Text Pruning (RTP) at training time. This setup focuses on pruning text from the captions to improve model robustness. The results indicate enhancements in various metrics compared to the hyperbolic baseline, such as 70.8 for TR@1, 95.2 for TR@10, 53.8 for IR@1, and 127.8 for CIDEr, but are still slightly lower than the Hyperbolic RQS model.

\paragraph{Hyperbolic RQS+RTP}
This configuration combines both Random Query selection (RQS) and Random Text Pruning (RTP) within the hyperbolic space. It achieves slighter lower performance on zero-shot retrieval compared to the Euclidean RQS+RTP counterpart, but significant improvements on the image captioning task, with scores of 79.3 (+0.5) for BLEU-1, 37.9 (+0.2) for BLEU-4, 31.0 (+1.1) for METEOR, 59.7 (+0.9) for ROUGE-L, 130.2 (+2.7) for CIDEr, and 23.3 (-0.1) for SPICE. These results highlight the combined effectiveness of RQS and RTP in enhancing the performance of the hyperbolic model.

%% file: figures/scripts/image_radii_hyper.tex
\begin{figure*}[ht]
\centering
\includegraphics[width=0.9\textwidth]{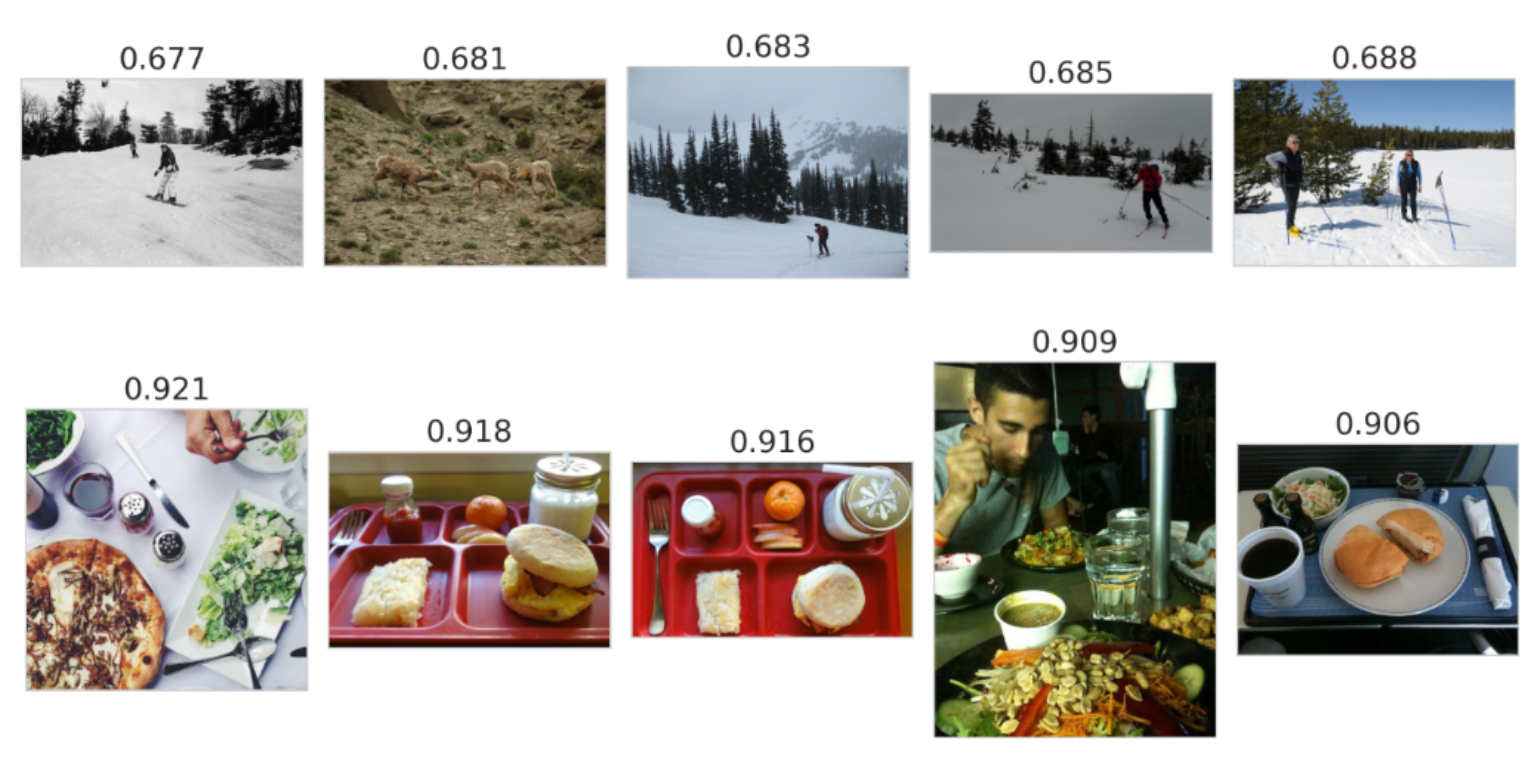}
\caption{(top-row) Images with lowest hyperbolic radius; (bottom-row) Images with highest hyperbolic radius. Radius is indicated above each image. 
}
\label{fig:image_radii_hyper}
\end{figure*}

%% file: figures/scripts/radii_distribution.tex
\begin{figure*}[t]
\centering
\includegraphics[width=1.\textwidth]{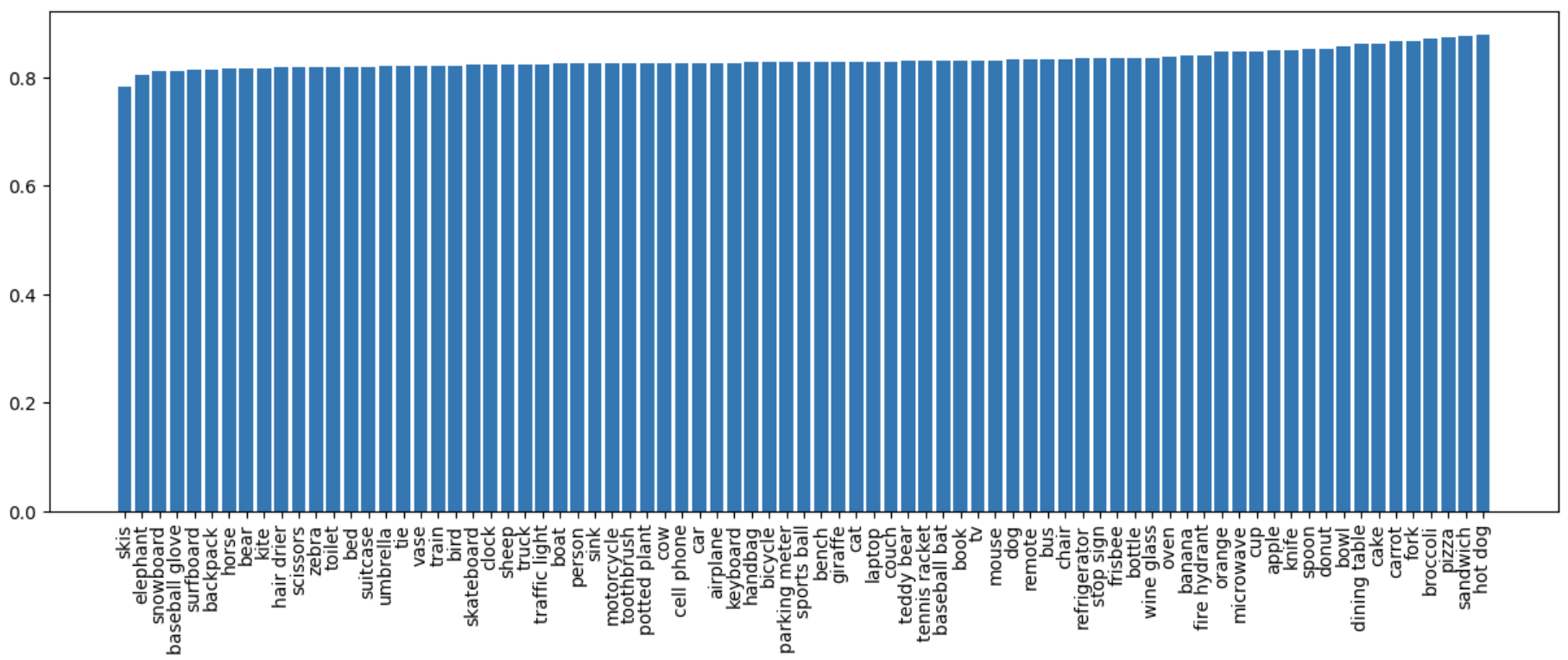}
\caption{Per class radius distribution, using the annotations in the COCO dataset. 
}
\label{fig:radii_distribution}
\end{figure*}

%% file: figures/scripts/hyper_radii_different_dimension.tex
\begin{figure*}[t]
\centering
\includegraphics[width=1.\textwidth]{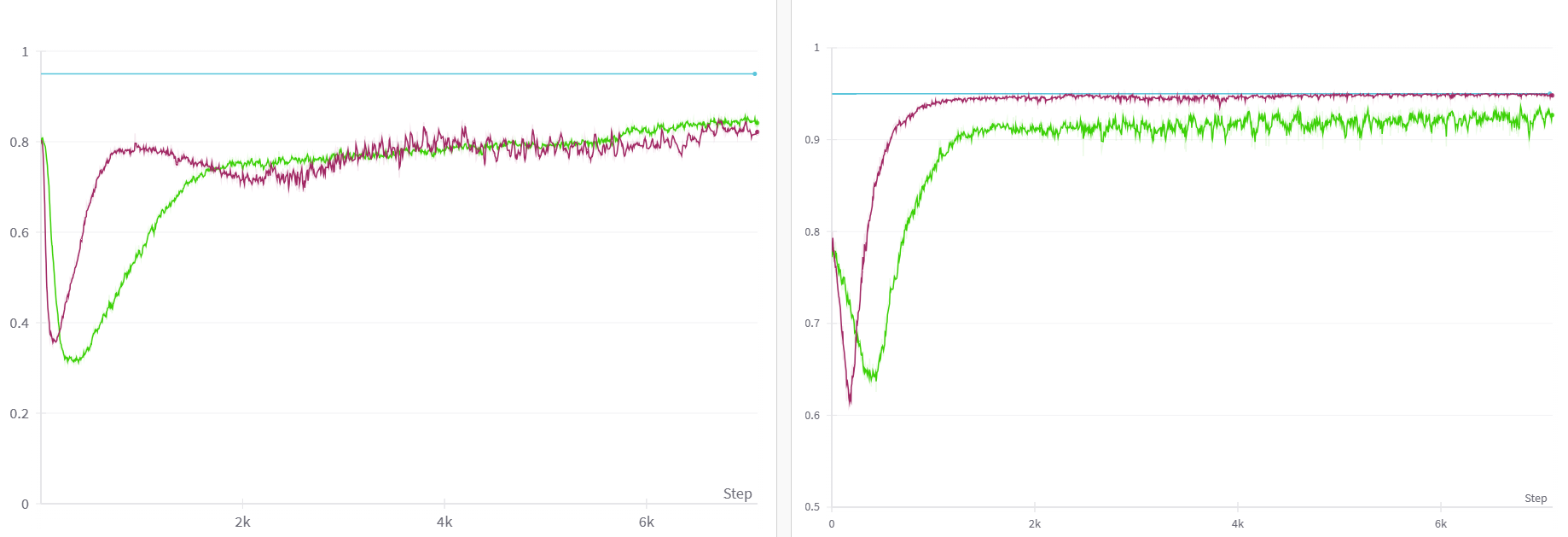}
\caption{Plot of average hyperbolic radius for the image embeddings (left) and text embeddings (right). The radius is stable at the maximum level using hidden dimension 256 (cyan), while it varies using a lower dimension, i.e., 16 (purple). The text embedding does not converge at the edge of the Poincaré ball, only using Random Text Pruning (green), resulting in a more variable and meaningful radius.
}
\label{fig:hyper_radii_different_dimension}
\vspace{-0.3cm}
\end{figure*}

%% file: figures/scripts/rand_query_plot.tex
\begin{figure*}[t]
\centering
\includegraphics[width=.9\textwidth]{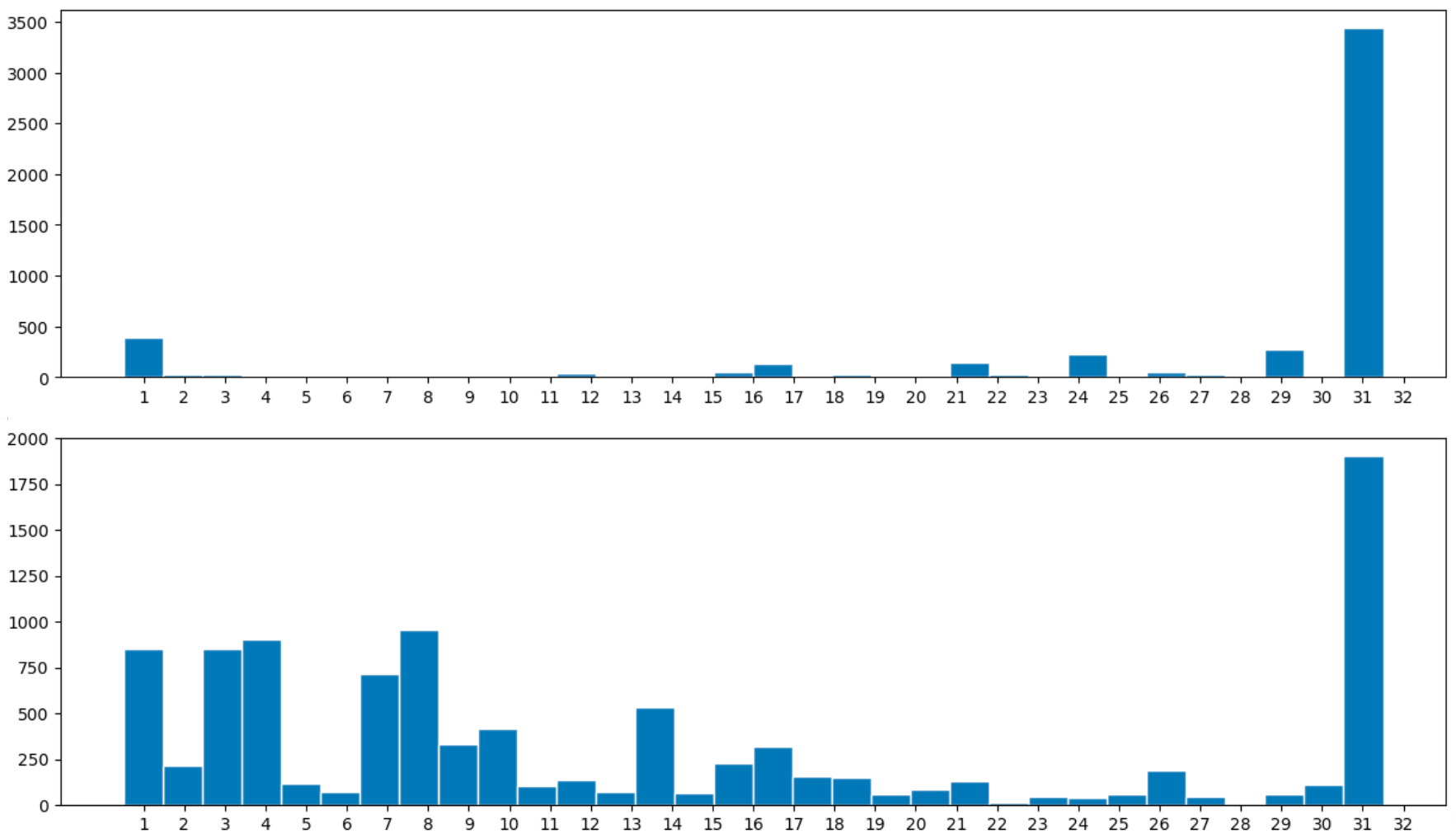}
\caption{Query selection distribution across the test dataset. The top graph depicts the selection performed by the BLIP-2 model, whereas the bottom graph shows the selection performed by our model. 
}
\label{fig:rand_query_plot}
\vspace{-0.2cm}
\end{figure*}

%% file: tables/main_results.tex
\begin{table}[h]
\centering
\resizebox{\textwidth}{!}{
\begin{tabular}{l|cccccc|cccccc}
\toprule
\multirow{2}{*}{Model} & \multicolumn{6}{c|}{COCO Fine-tuned (5K test set)} & \multicolumn{6}{c}{COCO Fine-tune Karpathy test} \\
 & TR@1 & TR@5 & TR@10 & IR@1 & IR@5 & IR@10 & B@1 & B@4 & M & R & C & S \\ \midrule
Euclidean baseline & 69.6 & 90.0 & 94.8 & 53.5 & 79.6 & 86.9 & 68.6 & 30.1 & 29.2 & 54.0 & 110.7 & 21.8 \\
Euclidean RQS & 72.1 & \textbf{92.1} & \textbf{96.2} & \textbf{55.6} & 80.9 & 87.8 & 77.3 & 36.8 & 29.2 & 57.9 & 127.5 & 23.1 \\
Euclidean RQS+RTP & \textbf{72.2} & \textbf{92.1} & 96.1 & \textbf{55.6} & \textbf{81.2} & \textbf{88.4} & 78.8 & 37.7 & 29.9 & 58.8 & 129.8 & \textbf{23.4} \\
Hyperbolic baseline & 67.0 & 88.0 & 91.5 & 51.2 & 77.5 & 85.0 & 65.0 & 28.5 & 28.0 & 52.5 & 105.0 & 20.5 \\
Hyperbolic (Poincaré) & 8.6 & 12.1 & 12.9 & 1.3 & 2.3 & 2.7 & 14.0 & 0.1 & 6.1 & 12.8 & 0.1 & 0.9 \\
Hyperbolic RQS & 71.3 & 91.4 & 95.8 & 54.2 & 80.7 & 87.1 & 77.5 & 37.1 & 28.7 & 58.1 & 128.8 & 23.2 \\
Hyperbolic RTP & 70.8 & 91.2 & 95.2 & 53.8 & 80.3 & 86.9 & 77.1 & 36.6 & 28.3 & 57.9 & 127.8 & 22.7 \\
Hyperbolic RQS+RTP & 71.9 & 91.6 & 96.0 & 54.5 & 80.2 & 87.7 & \textbf{79.3} & \textbf{37.9} & \textbf{31.0} & \textbf{59.7} & \textbf{130.2} & 23.3 \\
\bottomrule
\end{tabular}
}
\caption{Performance comparison of various models on COCO fine-tuned 5K test set for zero-shot image-text retrieval (TR) and image retrieval (IR), and on the COCO Karpathy test set for image captioning evaluated with BLEU (B), METEOR (M), ROUGE-L (R), CIDEr (C), and SPICE (S). Models include Euclidean and hyperbolic baselines, with and without Random Query Selection (RQS) and Random Text Pruning (RTP). All models use the ViT-g vision encoder and OPT 2.7B language model.}
\label{tab:main_results}
\end{table}

%% file: sections/discussion.tex
\section{Discussion}
\label{sec:discussion}

This study investigated the integration of hyperbolic embeddings in the large-scale BLIP-2 vision-language model, marking the first application of hyperbolic embeddings at this scale. While hyperbolic space theoretically offers advantages for capturing hierarchical relationships and uncertainty, we show that conventional approaches like Poincaré distance face different challenges.

Our hyperbolic training method achieved performance comparable or superior to Euclidean models while providing image embeddings with varying hyperbolic radii. Conventional hyperbolic approaches struggle both in performance and in capturing meaningful uncertainty. Our study demonstrates that it is possible to maintain or improve performance while incorporating a measure of uncertainty in the embeddings.

Moreover, the newly introduced Random Query Selection and Random Text Pruning techniques are independent of the embedding space used, despite being inspired by challenges faced when adapting BLIP-2 to hyperbolic space. Consequently, these techniques are broadly applicable and beneficial for Large Multimodal Models in general.

With our research, we want to highlight the limitations and the findings to advance the understanding of hyperbolic embeddings in VLMs. Future efforts should aim to refine hyperbolic methods to overcome scalability challenges and fully realize their potential in enhancing vision-language models.

%% file: sections/acknowledgements.tex
\section*{Acknowledgements}
\label{sec:ack}

The authors wish to thank Trevor Darrell for scientific discussions. We acknowledge financial support from the Panasonic Corporation, from the PNRR MUR project
PE0000013-FAIR and from the Sapienza grant RG123188B3EF6A80 (CENTS).